\documentclass[10pt,twocolumn,letterpaper]{article}

\usepackage{cvpr}              
\definecolor{cvprblue}{rgb}{0.21,0.49,0.74}
\definecolor{my_green}{RGB}{51,150,0}
\definecolor{my_yellow}{RGB}{255,165,0}
\definecolor{my_red}{RGB}{204, 0, 0}
\definecolor{my_green}{RGB}{51,152,0}
\definecolor{my_yellow}{RGB}{255,215,0}
\definecolor{my_red}{RGB}{204, 0, 0}
\definecolor{my_blue}{RGB}{0, 127, 254}
\definecolor{my_purple}{RGB}{127, 40, 244}
\definecolor{light_pink}{RGB}{250,245,247}
\definecolor{light_grey}{RGB}{240,240,240}
\definecolor{light_blue}{RGB}{60,65,215}
\definecolor{light_green}{RGB}{60,165,65}
\definecolor{light_red}{RGB}{204,0,0}
\definecolor{light_orange}{RGB}{210,105,30}

\usepackage[pagebackref,breaklinks,colorlinks,allcolors=cvprblue]{hyperref}
\renewcommand{\paragraph}[1]{\vspace{1.25mm}\noindent\textbf{#1}}
\usepackage{makecell}
\usepackage{pifont}
\renewcommand{\checkmark}{\textcolor{my_green}{\ding{51}}} 
\newcommand{\crossmark}{\textcolor{my_red}{\ding{55}}} 
\newcommand{\green}[1]{\textcolor{light_green}{#1}}
\newcommand{\pink}[1]{\textcolor{light_red}{#1}}
\newcommand{\blue}[1]{\textcolor{light_blue}{#1}}
\newcommand{\orange}[1]{\textcolor{light_orange}{#1}}
\newcommand{\myblue}[1]{\textcolor{my_blue}{#1}}

\usepackage{multirow}
\usepackage{graphicx}
\usepackage{arydshln}
\usepackage{setspace}

\usepackage{array}
\newcolumntype{x}[1]{>{\centering\arraybackslash}p{#1pt}}
\newcolumntype{y}[1]{>{\raggedright\arraybackslash}p{#1pt}}
\newcolumntype{z}[1]{>{\raggedleft\arraybackslash}p{#1pt}}

\usepackage{listings}
\usepackage{colortbl}

\lstdefinelanguage{json}{
  basicstyle=\ttfamily\fontsize{8pt}{10pt}\selectfont,
  numbers=left,
  numberstyle=\tiny\color{gray},
  stepnumber=1,
  numbersep=5pt,
  showstringspaces=false,
  breaklines=true,
  frame=single,
  backgroundcolor=\color{gray!10},
  literate=
   *{0}{{{\color{numb}0}}}{1}
    {1}{{{\color{numb}1}}}{1}
    {2}{{{\color{numb}2}}}{1}
    {3}{{{\color{numb}3}}}{1}
    {4}{{{\color{numb}4}}}{1}
    {5}{{{\color{numb}5}}}{1}
    {6}{{{\color{numb}6}}}{1}
    {7}{{{\color{numb}7}}}{1}
    {8}{{{\color{numb}8}}}{1}
    {9}{{{\color{numb}9}}}{1}
    {$m$}{{{\color{numb}m}}}{1}
    {$n$}{{{\color{numb}n}}}{1}
    {$n+1$}{{{\color{numb}{n+1}}}}{3}
    {$n+2$}{{{\color{numb}{n+2}}}}{3}
    {:}{{{\color{punct}{:}}}}{1}
    {,}{{{\color{punct}{,}}}}{1}
    {"}{{{\color{string}{"}}}}{1},
}

\definecolor{numb}{rgb}{0.6,0,0}
\definecolor{punct}{rgb}{0,0,0}
\definecolor{string}{rgb}{0,0.58,0}

\usepackage{minted}
\usepackage[most]{tcolorbox}
\def\paperID{*****} 
\def\confName{CVPR}
\def\confYear{2026}

\title{SpatiaLQA: A Benchmark for Evaluating Spatial Logical Reasoning in Vision-Language Models}

\author{
 Yuechen Xie\textsuperscript{\rm 1}\thanks{Equal Contribution, $^\dagger$Corresponding Author.}\;,
 Xiaoyan Zhang\textsuperscript{\rm 1}$^{*}$,
 Yicheng Shan\textsuperscript{\rm 2}$^{*}$,
 Zhu Hao\textsuperscript{\rm 3}, \\
 Rui Tang\textsuperscript{\rm 3},
 Rong Wei\textsuperscript{\rm 3},
 Mingli Song\textsuperscript{\rm 1,4,5},
 Yuanyu Wan\textsuperscript{\rm 1},
 Jie Song\textsuperscript{\rm 1}$^\dagger$ \\[2mm]
 $^1$Zhejiang University, $^2$The University of Sydney, $^3$ManyCore\\
 $^4$State Key Laboratory of Blockchain and Security, Zhejiang University \\
 $^5$Hangzhou High-Tech Zone (Binjiang) Institute of Blockchain and Data Security \\[2mm]
}

\begin{document}
\maketitle
\begin{abstract}
Vision-Language Models (VLMs) have been increasingly applied in real-world scenarios due to their outstanding understanding and reasoning capabilities. Although VLMs have already demonstrated impressive capabilities in common visual question answering and logical reasoning, they still lack the ability to make reasonable decisions in complex real-world environments. We define this ability as spatial logical reasoning, which not only requires understanding the spatial relationships among objects in complex scenes, but also the logical dependencies between steps in multi-step tasks. To bridge this gap, we introduce Spatial Logical Question Answering (SpatiaLQA), a benchmark designed to evaluate the spatial logical reasoning capabilities of VLMs. SpatiaLQA consists of 9,605 question answer pairs derived from 241 real-world indoor scenes. We conduct extensive experiments on 41 mainstream VLMs, and the results show that even the most advanced models still struggle with spatial logical reasoning. To address this issue, we propose a method called recursive scene graph assisted reasoning, which leverages visual foundation models to progressively decompose complex scenes into task-relevant scene graphs, thereby enhancing the spatial logical reasoning ability of VLMs, outperforming all previous methods. Code and dataset are available at \url{https://github.com/xieyc99/SpatiaLQA}.
\end{abstract}    
\section{Introduction}
\label{sec:intro}

\begin{figure*}[t]
  \centering
   \includegraphics[width=\linewidth, trim=0.7cm 0.2cm 0.8cm 0.2cm, clip]{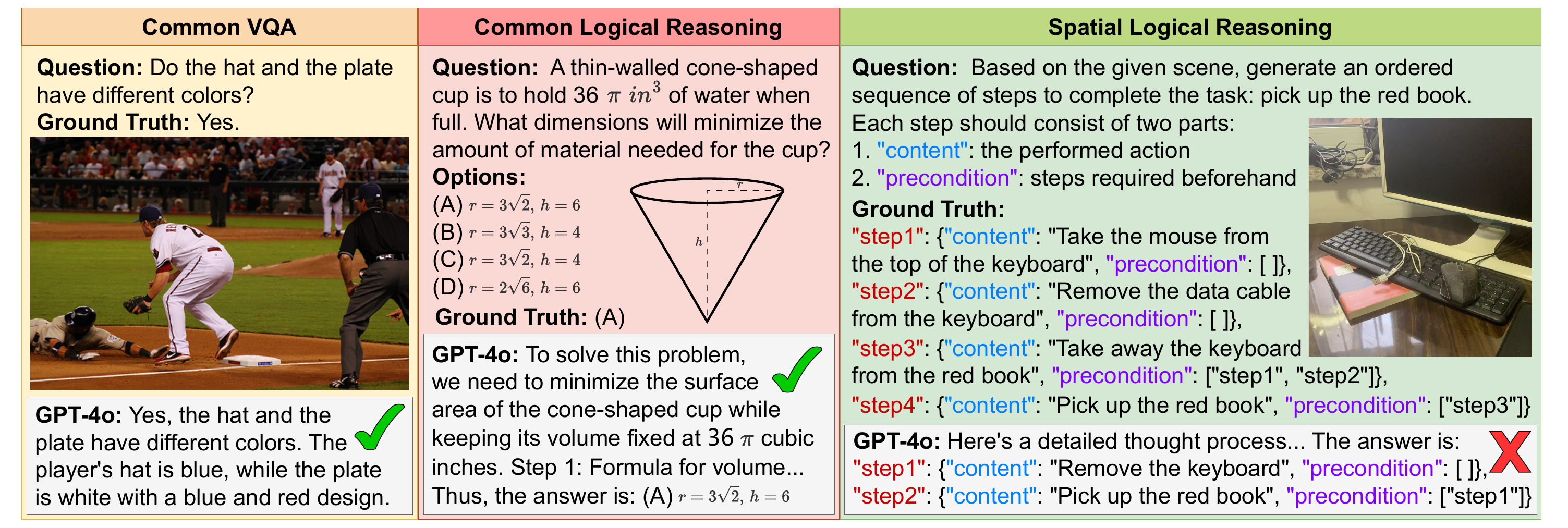}
   \vspace{-1.em}
   \caption{Common VQA~\cite{hudson2019gqa} typically involve recognizing visual content and factual knowledge, while common logical reasoning~\cite{yue2024mmmu} focuses on abstract, symbolic problem-solving. Spatial logical reasoning, in contrast, requires integrating both spatial understanding and multi-step logical reasoning to accomplish tasks in real-world scenes.}
   \vspace{-1em}
   \label{fig:motivation}
\end{figure*}

Vision-Language Models (VLMs)~\cite{li2022blip,li2023blip,team2025gemma,liu2024improved,liu2023visual,bai2025qwen2} have recently been increasingly applied to interpret and reason about complex real-world scenes, achieving remarkable progress across various domains such as Visual Question Answering (VQA)~\cite{antol2015vqa,marino2019ok,chow2025physbench,chen2024spatialvlm}, task planning~\cite{yang2025guiding,mei2024replanvlm,zhang2023grounding}, image captioning~\cite{yang2023exploring,luu2024questioning,xu2024pllava}, and scene understanding~\cite{liu2024vision,wang2024root,cao2024maplm,zhi2025lscenellm}. 
As shown in the first two examples of \cref{fig:motivation}, state-of-the-art VLMs such as GPT-4o~\cite{hurst2024gpt} perform well on common VQA~\cite{hudson2019gqa,singh2019towards,lin2022revive} and logical reasoning~\cite{yue2024mmmu,paglieri2024balrog} tasks. However, in the third example, it fails to remove the obstacles above the bottom book before picking it up, indicating that its performance on tasks requiring both spatial understanding and logical reasoning remains unsatisfactory. We refer to this important yet underexplored task as \textit{spatial logical reasoning}.
Such tasks not only require models to possess strong spatial understanding~\cite{cheng2024spatialrgpt,cai2025spatialbot} but also demand the ability to reason through a sequence of logically consistent steps~\cite{tong2025code2logic}.

Moreover, although spatial logical reasoning shares similarities with Embodied Question Answering (EQA)~\cite{li2024embodied,mei2024replanvlm,zhang2023grounding}, as both require an understanding of spatial relations and reasoning over multi-step dependencies, its research value is distinct and irreplaceable. The primary focus of EQA is to evaluate whether an agent can translate abstract language instructions into physically executable action sequences under the constraints of real-world dynamics and control strategies. These action sequences are typically selected from a predefined and limited set of motor primitives (\eg move forward, turn left, pick up), forming a closed output space. In contrast, spatial logical reasoning does not involve any execution component. Instead, it emphasizes whether the model can deduce a logically consistent and spatially coherent multi-step reasoning process purely at the visual-semantic level, where the answers belong to an open vocabulary space. This open-ended nature demands higher levels of cognitive and linguistic abstraction, reflecting a model’s intrinsic reasoning and compositional understanding rather than its ability to map instructions to preset fixed actions. Therefore, spatial logical reasoning serves as the cognitive basis for EQA, without relying on direct physical interactions. Advancing spatial logical reasoning is not only essential for improving performance in embodied tasks, but also for enhancing the overall reasoning capacity of VLMs across diverse real-world domains.
Unfortunately, existing benchmarks fail to systematically and accurately reflect the performance of VLMs in this aspect, leaving a critical gap that constrains their safe and effective deployment in real-world scenarios~\cite{zhao2025cot,sermanet2024robovqa,zhen20243d}.

To address this important yet unexplored issue, we introduce Spatial Logical Question Answering (SpatiaLQA) in this work, a benchmark dataset consisting of 9,605 image–text question answer (QA) pairs collected from 241 indoor scenes spanning 13 scene categories. Considering the difficulty of acquiring such data, particularly since scenes involving complex logical relationships need to be carefully arranged, the entire annotation process is divided into three stages: manual annotation, subgraph extraction augmentation, and graph expansion augmentation. Specifically, we first manually annotated 2,401 real indoor scene images, assigning one QA pair to each image. We then applied subgraph extraction augmentation to these 2,401 samples, which derives subsets of the original answer steps based on their logical dependencies, and obtained 2,251 new QA pairs. Finally, we performed graph expansion augmentation on the combined 4,652 samples, where heuristic methods were used to append several logically consistent steps to the original answers for data enrichment, resulting in 4,953 newly generated QA pairs.

In addition, we systematically evaluated 41 representative VLMs. Specifically, the evaluation process consists of three steps. First, we use GPT-4o to perform step-level matching between the model’s predictions and the ground-truth annotations. Next, the Hungarian algorithm~\cite{munkres1957algorithms} is applied to generate the optimal one-to-one step matching that achieves the maximum number of pairs. Finally, based on the matched results, we calculate precision and recall for both the content and the preconditions. Extensive experiments show that most current models perform poorly in spatial logical reasoning, especially in complex tasks that require many steps.

To improve the spatial logical reasoning capabilities of VLMs, we propose a method called \textit{recursive scene graph assisted reasoning}. Specifically, our method consists of three steps:
(1) We first use Depth Anything V2~\cite{yang2024depth} and SAM~\cite{kirillov2023segment} to obtain the depth map and segmentation map of the scene image;
(2) Based on the original image and these perception results, we take the object specified in the task as the initial \textit{source object} and perform the first round of scene graph generation using the VLM. This process identifies the objects in direct contact with the source object, referred to as \textit{target objects}, along with their relative spatial relationships. Then construct a scene graph with the source and target objects as nodes and the spatial relationships as edges. This scene graph serves as the input for the next iteration, where the previous target objects are regarded as new source objects, and the process repeats until reaching the maximum iteration number;
(3) Finally, the generated scene graph and the prompt are jointly fed into the VLM to produce the final answer. 
By leveraging domain-specialized visual foundation models, our method incrementally decomposes the complex visual scenes into task-relevant scene graphs. This hierarchical perception process allows the VLM to focus on the spatial environment surrounding the target objects, thereby facilitating more accurate multi-step reasoning.

\begin{table*}
\centering
\small
\addtolength{\tabcolsep}{-2.9pt}
\begin{tabular}{cccccccccc}
\Xhline{1.0pt}
\textbf{Benchmarks} & \textbf{Modality} & \textbf{SU} & \textbf{LR} & \textbf{Real Scene} & \textbf{Answer Type}  & \textbf{Multi-step} & \textbf{Precondition} & \textbf{LLM/VLM Scoring} & \textbf{Size} \\ \hline
\green{CLEVR}~\cite{johnson2017clevr} & I & \checkmark & \crossmark & \crossmark & Open & \crossmark & \crossmark & \crossmark & 853.6K \\
\green{GQA}~\cite{hudson2019gqa} & I & \checkmark & \crossmark & \checkmark & Open & \crossmark & \crossmark & \crossmark & $>$1M \\
\green{MMBench}~\cite{liu2024mmbench} & I & \checkmark & \crossmark & \checkmark & MC & \crossmark & \crossmark & \checkmark & 3.2K \\
\green{Spatial-MM}~\cite{shiri2024empirical} & I & \checkmark & \crossmark & \checkmark & MC & \crossmark & \crossmark & \crossmark & 2.3K \\
\green{SpatialRGPT-Bench}~\cite{cheng2024spatialrgpt} & I & \checkmark & \crossmark & \checkmark & MC & \crossmark & \crossmark & \crossmark & 1.5K \\
\green{Open3DVQA}~\cite{zhang2025open3dvqa} & I & \checkmark & \crossmark & \crossmark & Open & \crossmark & \crossmark & \checkmark & 9.0K \\
\orange{MathVista}~\cite{lu2023mathvista} & I & \crossmark & \checkmark & \crossmark & MC/Open & \crossmark & \crossmark & \crossmark & 6.1K \\
\orange{MMMU}~\cite{yue2024mmmu} & I & \crossmark & \checkmark & \crossmark & MC/Open & \crossmark & \crossmark & \crossmark & 11.5K \\
\orange{GeoQA}~\cite{chen2021geoqa} & I & \crossmark & \checkmark & \crossmark & MC & \crossmark & \crossmark & \crossmark & 5.0K \\
\orange{ChartQA}~\cite{masry2022chartqa} & I & \crossmark & \checkmark & \crossmark & Open & \crossmark & \crossmark & \crossmark & 32.7K \\
\blue{MT-EQA}~\cite{yu2019multi} & I & \checkmark & \checkmark & \crossmark & MC & \crossmark & \crossmark & \crossmark & 20.0K \\
\blue{MP3D-EQA}~\cite{wijmans2019embodied} & I/P & \checkmark & \checkmark & \checkmark & MC & \crossmark & \crossmark & \crossmark & 1.1K \\
\blue{OpenEQA}~\cite{majumdar2024openeqa} & I/V & \checkmark & \checkmark & \checkmark & Open & \crossmark & \crossmark & \checkmark & 2.1K \\
\blue{EmbSpatial-Bench}~\cite{du2024embspatial} & I & \checkmark & \checkmark & \checkmark & MC & \crossmark & \crossmark & \crossmark & 3.6K \\
\blue{EmbodiedBench}~\cite{yang2025embodiedbench} & I & \checkmark & \checkmark & \crossmark & MC & \checkmark & \crossmark & \crossmark & 1.1K \\
\hline
SpatiaLQA (Ours) & I & \checkmark & \checkmark & \checkmark & Open & \checkmark & \checkmark & \checkmark & 9.6K \\
\Xhline{1.0pt}
\end{tabular}
\caption{Comparison between SpatiaLQA and other benchmarks (\green{VQA}, \orange{logical reasoning} and \blue{EQA}). `SU' and `LR' denote spatial understanding and long-range reasoning, respectively. `I', `P' and `V' denote image, point cloud and video, respectively. `Open' and `MC' stand for open-vocabulary and multiple-choice respectively. `Multi-step' specifies whether the answers involve multiple steps. `Precondition' indicates whether each step in the answer is annotated with its preconditions (i.e., which steps must be completed beforehand).}
\label{tab:related}
\vspace{-1em}
\end{table*}

In this work, we make four primary contributions:
(1) We identify and define spatial logical reasoning as a critical yet underexplored capability of VLMs, highlighting its importance for reasoning across interdependent spatial and logical steps in real-world scenarios.
(2) We introduce SpatiaLQA, a large-scale benchmark consisting of 9,605 image–text QA pairs across 241 indoor scenes spanning 13 scene categories, to comprehensively evaluate spatial logical reasoning.
(3) We conduct a systematic evaluation of 41 representative VLMs using GPT-4o and the Hungarian algorithm, revealing that most models struggle with spatial logical reasoning, particularly in complex tasks that require many steps.
(4) We propose a novel method, recursive scene graph assisted reasoning, which utilizes visual foundation models to decompose complex scenes into task-specific scene graphs, improving the spatial logical reasoning capability of VLMs.
\section{Related Work}
\label{sec:related}

We present the main differences between SpatiaLQA and some representative benchmarks in \cref{tab:related}, and provide detailed analyses of how SpatiaLQA differs from \green{VQA}, \orange{logical reasoning}, and \blue{EQA} in \cref{sec:benchmarks}. \cref{sec:vlm} presents the current state of VLMs.

\subsection{Related Benchmarks}
\label{sec:benchmarks}

\paragraph{Visual Question Answering.} Common VQA~\cite{antol2015vqa,malinowski2014multi,marino2019ok} primarily focused on recognizing image content and handling short-range reasoning tasks, such as object or attribute recognition~\cite{hudson2019gqa}, counting~\cite{johnson2017clevr}, and simple spatial relation reasoning~\cite{majumdar2024openeqa}. In addition, their efforts largely remained at the level of single-step factual question answering. In contrast, SpatiaLQA emphasizes spatial logical reasoning, where the model must perform long-range reasoning and infer a sequence of dependent operations based on spatial relations, rather than simply outputting a factual answer.

\paragraph{Logical Reasoning.} Logical reasoning tasks, such as mathematical reasoning~\cite{yue2024mmmu,lu2023mathvista,chen2021geoqa}, assess a model's ability to establish causal, implicational, and consistency relationships within complex information. However, these tasks are mostly confined to abstract textual or symbolic spaces, where the reasoning process is decoupled from real-world spatial information and visual structures. In contrast, SpatiaLQA focuses on spatial logical reasoning, which requires models to jointly understand spatial relationships and causal logic within realistic scenes. Fundamentally, spatial logical reasoning bridges spatial understanding and logical reasoning, representing a more challenging and practically significant form of reasoning.

\paragraph{Embodied Question Answering.} EQA focuses on the feasibility and effectiveness of actions in interactive environments, such as navigation~\cite{thomason2018shifting,zhong2025robotrom,majumdar2024openeqa} and manipulation~\cite{deng2020mqa,mei2024replanvlm,li2024embodied} tasks. The primary focus of EQA is to evaluate whether an agent can translate abstract language instructions into physically executable action sequences under the constraints of real-world dynamics and control strategies. These action sequences are typically selected from a predefined and limited set of motor primitives, forming a closed output space. In contrast, SpatiaLQA emphasizes whether the model can deduce a logically consistent and spatially coherent multi-step reasoning process purely at the visual-semantic level, where the answers belong to an open vocabulary space. This reasoning ability, namely spatial logical reasoning, forms the cognitive foundation for embodied tasks without relying on physical interaction.

\begin{figure*}[t]
  \centering
   \includegraphics[width=\linewidth, trim=0.7cm 0.0cm 0.7cm 0.0cm, clip]{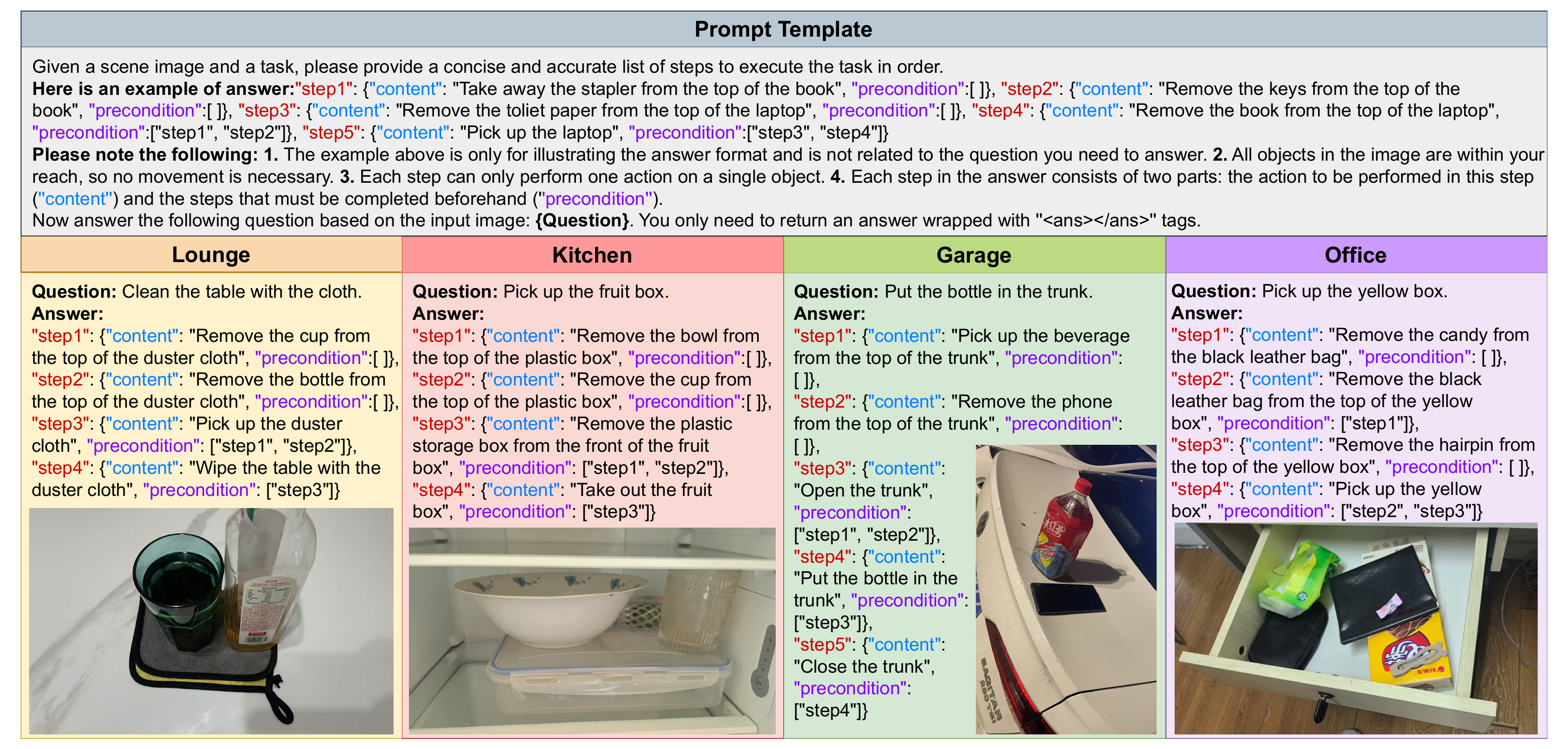}
   \vspace{-2.0em}
   \caption{Prompt template and examples of several indoor scenes.}
   \vspace{-0.5em}
   \label{fig:examples}
\end{figure*}

\begin{figure*}[t]
  \centering
   \includegraphics[width=\linewidth, trim=1.8cm 12.9cm 2.1cm 2.4cm, clip]{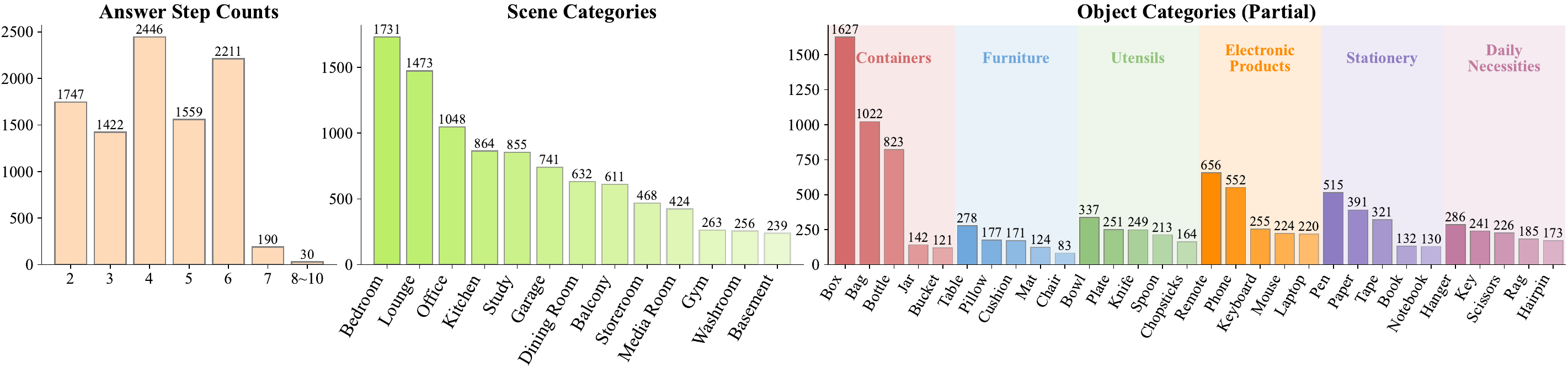}
   \vspace{-2.0em}
   \caption{The distributions of answer step counts, scene categories, and partial object categories in SpatiaLQA. The x-axes of the three plots represent the number of answer steps, indoor scene categories, and object categories, while the y-axes indicate the number of samples.}
   \vspace{-1.0em}
   \label{fig:overview_rect}
\end{figure*}

\subsection{Vision-Language Models}
\label{sec:vlm}

Recently, VLMs have achieved remarkable success in tasks such as image captioning~\cite{yang2023exploring,luu2024questioning,xu2024pllava}, visual question answering~\cite{antol2015vqa,malinowski2014multi}, and open-ended multimodal dialogue~\cite{ji2024wavchat,xue2025mmrc}, driven by large-scale pretraining~\cite{lin2024vila,wang2024cogvlm}, instruction tuning~\cite{chen2024your,ko2025st}, and external tool augmentation~\cite{qi2024rora}. Despite their strong generalization and reasoning capabilities, their performance in complex scenarios that require reasoning across multiple interdependent steps has not been systematically studied. SpatiaLQA directly addresses this gap by requiring VLMs to perform ordered and dependency-aware spatial logical reasoning under structured scene constraints. It further provides a systematic analysis of model capabilities in two key aspects: the generation of step content and the inference of preconditions, which enables a comprehensive evaluation of the VLMs' reasoning consistency and spatial understanding, thereby supporting their reliable and safe deployment in real-world scenarios.
\section{SpatiaLQA}
\label{sec:spatialqa}

To evaluate the spatial logical reasoning capabilities of VLMs, we first define the concept of spatial logical reasoning and introduce the SpatiaLQA dataset in \cref{overview}. Then, we describe the dataset collection process and evaluation metrics in \cref{collection} and \cref{metric}, respectively. In \cref{evaluation}, we conduct experiments using SpatiaLQA to determine whether VLMs can effectively perform spatial logical reasoning. Finally, in \cref{analysis}, we analyze the alignment between our metrics and human evaluations, as well as the potential reasons for the poor performance of VLMs.

\begin{figure*}[t]
  \centering
   \includegraphics[width=\linewidth, trim=0.7cm 0.1cm 0.7cm 0.3cm, clip]{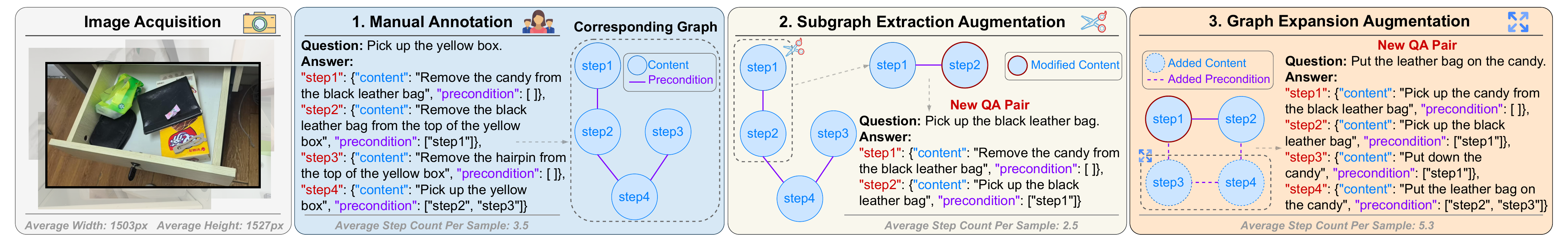}
   \vspace{-2.0em}
   \caption{The data collection pipeline for SpatiaLQA. Note that although the graph expansion augmentation in the figure is applied only to the data from subgraph extraction augmentation, we actually also applied graph expansion augmentation to the manually annotated data.}
   \vspace{-1.0em}
   \label{fig:data_collection}
\end{figure*}

\subsection{Overview of SpatiaLQA}
\label{overview}

Spatial logical reasoning refers to the ability of a model to solve complex problems by outputting a series of logically coherent steps through spatial understanding and logical reasoning. This capability is crucial yet fundamentally challenging for VLMs, as the model must integrate spatial understanding with logical reasoning, which involves precise spatial perception and tightly coordinated multi-step causal reasoning to ensure safe and effective operation.

However, existing datasets often focus solely on either spatial understanding or logical reasoning, while neglecting the integrated aspect of spatial logical reasoning described above. To bridge this gap, we introduce SpatiaLQA, a benchmark designed to comprehensively evaluate the spatial logical reasoning capabilities of VLMs. The dataset consists of 9,605 QA pairs collected from 241 scenes across 13 real-world indoor scene categories. The prompt template and QA examples are shown in \cref{fig:examples}, which includes the required answer format and example outputs that the model should follow. \cref{fig:overview_rect} shows the distribution of answer step counts, scene categories, and partial object categories. It shows that the answer step counts are broadly distributed across the range of 2–10, reflecting a diverse level of task complexity (in general, questions with more answer steps are more challenging). Additionally, the dataset’s images come from 13 common indoor scene categories, and the QA pairs involve over a thousand distinct objects (only a subset is shown in the figure), which demonstrates that SpatiaLQA encompasses a rich variety of scenes and objects.

\subsection{Dataset Collection Process}
\label{collection}

We first collected 2,401 real indoor scene images from 241 locations across 13 scene categories, each depicting a complex, multi-step task.
Given the challenges in collecting such data, particularly because scenes with complex logical dependencies require deliberate setup, as illustrated in \cref{fig:data_collection}, the collection process consists of three stages: manual annotation, subgraph extraction augmentation, and graph expansion augmentation. The details are as follows:

\paragraph{Manual Annotation.} We first employed trained annotators to manually label the 2,401 images, assigning one QA pair per image with answers ranging from 2–8 steps. Notably, we did not annotate implicit preconditions: for instance, if `step$k$' depends on `step$j$' and `step$l$' depends on `step$k$', we do not mark `step$j$' as a precondition of `step$l$'.
To ensure data quality, we conducted two rounds of review and correction by professional annotators. Each annotation was then represented as an undirected graph, where nodes correspond to step contents and edges indicate logical dependencies between steps, serving as the basis for the following augmentation stages.

\paragraph{Subgraph Extraction Augmentation.} We applied subgraph extraction augmentation to these 2,401 samples, generating 2,251 new QA pairs. This method derives subgraphs (each containing at least one edge) from the original annotations based on their logical dependencies, forming new QA pairs that share the same image, with the question corresponding to the final step of the subgraph.

\paragraph{Graph Expansion Augmentation.} We generated 4,953 new QA pairs using graph expansion augmentation, based on the samples containing `Remove' and `Pick up' from the previous two stages. Specifically, assuming the original question and the final step is `Pick up B', with an intermediate step being `Remove A', the graph expansion augmentation would change the question to `Place B on A', modify `Remove A' to `Pick up A', and add two additional steps at the end: `Put down A' and `Place B on A'. The generated QA pairs share the same image as the original sample.


\subsection{Evaluation Metrics}
\label{metric}

\begin{figure}[t]
  \centering
   \includegraphics[width=\linewidth, trim=0.7cm 0.2cm 1.0cm 0.2cm, clip]{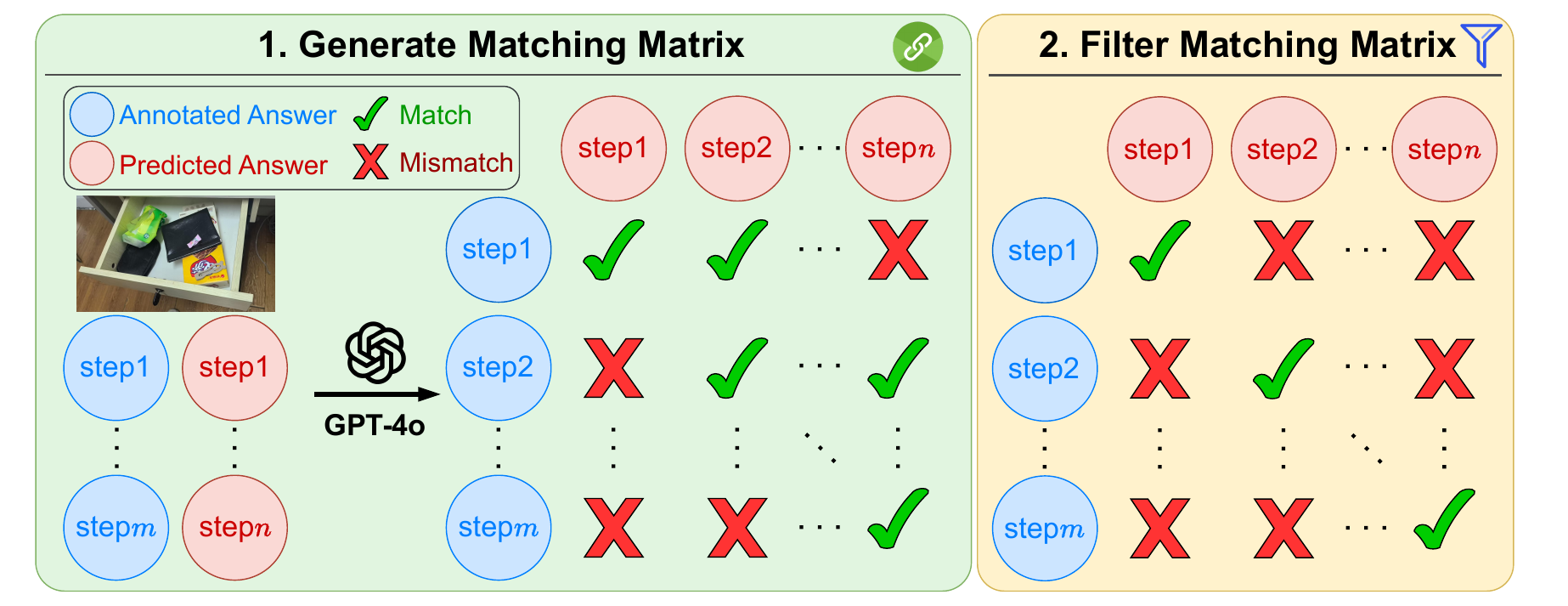}
   \vspace{-2.0em}
   \caption{The matching process between the predicted and annotated steps. We first use GPT-4o to match the predicted steps and annotated steps in pairs based on the image (allowing one-to-many matches), resulting in a matching matrix. Then, we apply the Hungarian algorithm to filter the matching matrix, removing redundant matches to achieve the maximum one-to-one matches.}
   \vspace{-1.0em}
   \label{fig:metric}
\end{figure}

Although human evaluation is the gold standard for open-vocabulary tasks, it is costly and time-consuming, making automatic metrics preferable for benchmarking. To this end, we first use GPT-4o and the Hungarian algorithm to match the predicted results with the ground-truth annotations, as shown in \cref{fig:metric}, and then calculate the recall and precision based on the matching results. Specifically, the evaluation process is divided into three steps:
(1) Use GPT-4o to match the predicted steps with the ground truth steps based on the image, i.e., determining whether the semantics of the content in each step are consistent in the image. Given an annotated answer with $m$ steps and a predicted answer with $n$ steps, we use GPT-4o to generate a matching matrix with $m$ rows and $n$ columns, where the values are either 0 or 1. 0 means the two steps differ, while 1 means they are the same. Note that in this step, a predicted (annotated) step can be matched with multiple annotated (predicted) steps.
(2) Apply the Hungarian algorithm to filter the matching matrix, removing redundant matches and achieving the maximum one-to-one match between the predicted steps and the annotated steps, resulting in the filtered matching matrix.
(3) Using the filtered matching matrix for all samples, we calculated the recall $R_c$ and precision $P_c$ for the all content, as well as the recall $R_p$ and precision $P_p$ for the all preconditions. Finally, we used the F1 score $F_c$ and $F_p$ for content and preconditions as the evaluation metrics.

Please refer to the supplementary material for the prompt used to generate the matching matrix.

\subsection{The Evaluation of VLMs}
\label{evaluation}

To evaluate the spatial logical reasoning capabilities of VLMs, we conducted experiments on SpatiaLQA with 41 representative VLMs, covering a wide range of types, including those in non-thinking mode and thinking mode, as well as VLMs that only support text–image input and general VLMs that support mixed multimodal inputs.
For details on VLM prompts and hyperparameters, please refer to the supplementary materials.

\begin{table}
\centering
\footnotesize
\addtolength{\tabcolsep}{-3.2pt}
\begin{tabular}{cccccccc}
\Xhline{1.0pt}
& \textbf{Size} & $R_c$ & $P_c$ & $F_c$ & $R_p$ & $P_p$ & $F_p$ \\ \hline
human & - & \textcolor{gray!80}{97.6} & \textcolor{gray!80}{97.6} & 97.6 & \textcolor{gray!80}{92.3} & \textcolor{gray!80}{92.7} & 92.5 \\ \hline
BLIP2-OPT~\cite{li2023blip} & 4B & \textcolor{gray!80}{24.6} & \textcolor{gray!80}{99.9} & 39.5 & \textcolor{gray!80}{0.0} & \textcolor{gray!80}{0.0} & -  \\
BLIP2-Flan-T5-xl~\cite{li2023blip} & 4B & \textcolor{gray!80}{24.0} & \textcolor{gray!80}{74.3} & 36.3 & \textcolor{gray!80}{0.1} & \textcolor{gray!80}{0.2} & 0.1  \\
BLIP2-Flan-T5-xxl~\cite{li2023blip} & 12B & \textcolor{gray!80}{41.3} & \textcolor{gray!80}{67.7} & 51.3 & \textcolor{gray!80}{0.0} & \textcolor{gray!80}{0.2} & 0.0  \\
LLaVA1.5-7B~\cite{liu2023visual} & 7B & \textcolor{gray!80}{40.3} & \textcolor{gray!80}{39.8} & 40.0 & \textcolor{gray!80}{6.4} & \textcolor{gray!80}{9.0} & 7.5  \\
LLaVA1.5-13B~\cite{liu2023visual} & 13B & \textcolor{gray!80}{39.1} & \textcolor{gray!80}{49.4} & 42.5 & \textcolor{gray!80}{9.1} & \textcolor{gray!80}{19.3} & 12.3  \\
LLaVA1.6-mistral~\cite{liu2024llavanext} & 7B & \textcolor{gray!80}{41.5} & \textcolor{gray!80}{47.3} & 44.2 & \textcolor{gray!80}{7.2} & \textcolor{gray!80}{13.6} & 9.4  \\
LLaVA1.6-vicuna-7B~\cite{liu2024llavanext} & 7B & \textcolor{gray!80}{47.0} & \textcolor{gray!80}{48.6} & 47.8 & \textcolor{gray!80}{10.0} & \textcolor{gray!80}{15.5} & 12.2  \\
LLaVA1.6-vicuna-13B~\cite{liu2024llavanext} & 13B & \textcolor{gray!80}{42.0} & \textcolor{gray!80}{48.5} & 45.0 & \textcolor{gray!80}{11.9} & \textcolor{gray!80}{24.9} & 16.1  \\
Phi3.5-vision-Ins~\cite{abdin2024phi} & 4B & \textcolor{gray!80}{36.6} & \textcolor{gray!80}{38.4} & 37.5 & \textcolor{gray!80}{5.5} & \textcolor{gray!80}{10.3} & 7.2  \\
Gemma3-12B-Ins~\cite{team2025gemma} & 12B & \textcolor{gray!80}{41.0} & \textcolor{gray!80}{64.6} & 50.2 & \textcolor{gray!80}{9.1} & \textcolor{gray!80}{24.9} & 13.4  \\
Gemma3-27B-Ins~\cite{team2025gemma} & 27B & \textcolor{gray!80}{44.1} & \textcolor{gray!80}{60.5} & 51.0 & \textcolor{gray!80}{9.9} & \textcolor{gray!80}{20.6} & 13.4  \\
Qwen2.5-VL-3B-Ins~\cite{bai2025qwen2} & 3B & \textcolor{gray!80}{32.5} & \textcolor{gray!80}{75.8} & 45.5 & \textcolor{gray!80}{3.3} & \textcolor{gray!80}{7.4} & 4.6  \\
Qwen2.5-VL-7B-Ins~\cite{bai2025qwen2} & 7B & \textcolor{gray!80}{32.5} & \textcolor{gray!80}{73.2} & 45.1 & \textcolor{gray!80}{3.3} & \textcolor{gray!80}{15.5} & 5.5  \\
Qwen2.5-VL-32B-Ins~\cite{bai2025qwen2} & 32B & \textcolor{gray!80}{42.0} & \textcolor{gray!80}{75.5} & 54.0 & \textcolor{gray!80}{9.2} & \textcolor{gray!80}{24.6} & 13.4  \\
Qwen2.5-VL-72B-Ins~\cite{bai2025qwen2} & 72B & \textcolor{gray!80}{60.0} & \textcolor{gray!80}{82.9} & 69.6 & \textcolor{gray!80}{23.9} & \textcolor{gray!80}{44.8} & 31.2  \\
Qwen3-VL-4B-Ins~\cite{team2025qwen3} & 4B & \textcolor{gray!80}{38.8} & \textcolor{gray!80}{78.8} & 52.0 & \textcolor{gray!80}{12.5} & \textcolor{gray!80}{46.5} & 19.6  \\
Qwen3-VL-8B-Ins~\cite{team2025qwen3} & 8B & \textcolor{gray!80}{44.0} & \textcolor{gray!80}{65.7} & 52.7 & \textcolor{gray!80}{13.4} & \textcolor{gray!80}{36.2} & 19.0  \\
Cosmos-Reason1~\cite{azzolini2025cosmos} & 7B & \textcolor{gray!80}{48.2} & \textcolor{gray!80}{66.9} & 56.1 & \textcolor{gray!80}{11.9} & \textcolor{gray!80}{40.3} & 18.3  \\
InternVL3.5-4B-Ins~\cite{wang2025internvl3_5} & 4B & \textcolor{gray!80}{42.0} & \textcolor{gray!80}{67.8} & 51.9 & \textcolor{gray!80}{11.3} & \textcolor{gray!80}{30.5} & 16.5  \\
InternVL3.5-8B-Ins~\cite{wang2025internvl3_5} & 8B & \textcolor{gray!80}{43.1} & \textcolor{gray!80}{71.5} & 53.8 & \textcolor{gray!80}{13.4} & \textcolor{gray!80}{33.6} & 19.2  \\
InternVL3.5-14B-Ins~\cite{wang2025internvl3_5} & 14B & \textcolor{gray!80}{49.3} & \textcolor{gray!80}{66.8} & 56.8 & \textcolor{gray!80}{14.0} & \textcolor{gray!80}{24.9} & 17.9  \\
GLM-4.1V-9B-Base~\cite{hong2025glm} & 9B & \textcolor{gray!80}{45.6} & \textcolor{gray!80}{78.1} & 57.6 & \textcolor{gray!80}{12.1} & \textcolor{gray!80}{31.2} & 17.5  \\
GLM-4.1V-9B-Thinking~\cite{hong2025glm} & 9B & \textcolor{gray!80}{44.0} & \textcolor{gray!80}{82.7} & 57.5 & \textcolor{gray!80}{12.4} & \textcolor{gray!80}{37.1} & 18.6  \\
Kimi-VL-A3B-Ins~\cite{team2025kimi} & 16B & \textcolor{gray!80}{28.7} & \textcolor{gray!80}{78.2} & 42.0 & \textcolor{gray!80}{1.1} & \textcolor{gray!80}{11.4} & 2.1  \\
Kimi-VL-A3B-Thinking~\cite{team2025kimi} & 16B & \textcolor{gray!80}{32.0} & \textcolor{gray!80}{92.2} & 47.5 & \textcolor{gray!80}{4.4} & \textcolor{gray!80}{42.3} & 8.0  \\
DeepSeek-VL2-Small~\cite{wu2024deepseek} & 16B & \textcolor{gray!80}{43.4} & \textcolor{gray!80}{84.9} & 57.4 & \textcolor{gray!80}{10.5} & \textcolor{gray!80}{59.1} & 17.9  \\
MiniCPM-V-4.5~\cite{yu2025minicpmv45cookingefficient} & 9B & \textcolor{gray!80}{50.0} & \textcolor{gray!80}{59.4} & 54.2 & \textcolor{gray!80}{10.9} & \textcolor{gray!80}{18.6} & 13.7  \\
SpaceOm~\cite{chen2024spatialvlm} & 4B & \textcolor{gray!80}{34.1} & \textcolor{gray!80}{71.2} & 46.1 & \textcolor{gray!80}{8.7} & \textcolor{gray!80}{35.1} & 13.9  \\
Pixtral~\cite{agrawal2024pixtral} & 12B & \textcolor{gray!80}{48.1} & \textcolor{gray!80}{70.0} & 57.0 & \textcolor{gray!80}{13.7} & \textcolor{gray!80}{31.2} & 19.0  \\ \hdashline[3pt/1.5pt]
 Qwen-VL-Plus~\cite{bai2023qwen} &  - &  \textcolor{gray}{38.3} &  \textcolor{gray!80}{86.0} &  53.0 &  \textcolor{gray!80}{9.5} &  \textcolor{gray!80}{37.5} &  15.1  \\
 Qwen-VL-Max~\cite{bai2023qwen} &  - &  \textcolor{gray!80}{62.0} &  \textcolor{gray!80}{83.0} &  70.9 &  \textcolor{gray!80}{25.6} &  \textcolor{gray!80}{45.2} &  32.7  \\
 GPT-4o-mini~\cite{hurst2024gpt} &  - &  \textcolor{gray!80}{54.3} &  \textcolor{gray!80}{76.4} &  63.5 &  \textcolor{gray!80}{14.1} &  \textcolor{gray!80}{26.4} &  18.4  \\
 GPT-4o~\cite{hurst2024gpt} &  - &  \textcolor{gray!80}{59.0} &  \textcolor{gray!80}{78.5} &  67.4 &  \textcolor{gray!80}{19.0} &  \textcolor{gray!80}{37.0} &  25.1  \\
 GPT-4.1-mini~\cite{openai_gpt41} &  - &  \textcolor{gray!80}{53.8} &  \textcolor{gray!80}{87.1} &  66.5 &  \textcolor{gray!80}{21.2} &  \textcolor{gray!80}{51.4} &  30.0  \\
 GPT-4.1~\cite{openai_gpt41} &  - &  \textcolor{gray!80}{65.2} &  \textcolor{gray!80}{84.2} &  73.5 &  \textcolor{gray!80}{30.2} &  \textcolor{gray!80}{51.3} &  38.0  \\
 GPT-5-mini~\cite{openai_gpt5_systemcard} &  - &  \textcolor{gray!80}{64.5} &  \textcolor{gray!80}{86.6} &  73.9 &  \textcolor{gray!80}{32.8} &  \textcolor{gray!80}{56.2} &  \myblue{41.2}  \\
 GPT-5~\cite{openai_gpt5_systemcard} &  - &  \textcolor{gray!80}{67.8} &  \textcolor{gray!80}{86.4} &  \pink{76.0} &  \textcolor{gray!80}{39.2} &  \textcolor{gray!80}{58.5} &  \pink{47.0}  \\
 Claude-3-7-sonnet~\cite{anthropic20253} &  - &  \textcolor{gray!80}{47.1} &  \textcolor{gray!80}{80.2} &  59.3 &  \textcolor{gray!80}{19.0} &  \textcolor{gray!80}{52.3} &  27.9  \\
 Claude-4-sonnet~\cite{anthropic20255} &  - &  \textcolor{gray!80}{59.4} &  \textcolor{gray!80}{82.0} &  68.9 &  \textcolor{gray!80}{27.8} &  \textcolor{gray!80}{\textcolor{gray!80}{52.1}} &  36.3  \\
 Gemini-2.5-flash~\cite{comanici2025gemini} &  - &  \textcolor{gray!80}{62.0} &  \textcolor{gray!80}{88.6} &  72.9 &  \textcolor{gray!80}{29.5} &  \textcolor{gray!80}{57.3} &  38.9  \\
 Gemini-2.5-pro~\cite{comanici2025gemini} &  - &  \textcolor{gray!80}{65.3} &  \textcolor{gray!80}{86.2} &  \myblue{74.3} &  \textcolor{gray!80}{31.0} &  \textcolor{gray!80}{53.6} &  39.3  \\
\Xhline{1.0pt}
\end{tabular}

\vspace{-1em}
\caption{The evaluation results of 41 VLMs. `Ins' indicates that the model is an `instruction-tuned' version. Recall and precision are used as reference metrics and are marked in \textcolor{gray!80}{gray}. The best and second-best F1 scores (excluding human results) are marked in \pink{red} and \myblue{blue}, respectively, and we use a dashed line to separate open-source VLMs (above) and proprietary VLMs (below).}
\label{tab:eval}
\vspace{-1.5em}
\end{table} 

In addition, we recruited a human participant to establish human-level performance on SpatiaLQA. We provided the participant with an answer template in \cref{fig:examples} and asked him to sequentially answer all the questions in SpatiaLQA.

The evaluation results are shown in \cref{tab:eval}, with the F1 scores being our primary metrics. We first share some observations and comments as follows:

(1) Generally, within the same series, VLMs with larger parameter sizes perform better, newer versions of VLMs perform better (\eg, Qwen2.5-VL-7B-Ins vs. Qwen3-VL-4B-Ins), and VLMs with a thinking mode outperform those with a non-thinking mode (\eg, Kimi-VL-A3B-Ins vs. Kimi-VL-A3B-Thinking). In addition, the performance of proprietary models typically surpasses that of open-source models. These confirm the effectiveness of the benchmarking and the correctness of the evaluation metrics.

(2) Humans achieved excellent performance on the benchmark, with the lowest metric exceeding 90\%. However, even the best-performing VLMs show a significant gap compared to humans, particularly in precondition prediction, which indicates that there is a notable performance gap between VLMs and humans in spatial logical reasoning.

(3) The prediction of preconditions generally performs worse than the prediction of content, indicating that VLMs have significant deficiencies in causal reasoning. Even though they may predict the steps roughly, they do not understand the logical relationships between them.

(4) The recall for content and preconditions is generally lower than precision, indicating that VLMs tend to output more certain answers during prediction, avoiding incorrect `false positives'. Specifically, the model generates very certain step contents or preconditions, but for uncertain steps, it may choose not to predict or skip them, leading to the omission of some steps that should have been identified. According to the statistics, the best-performing GPT-5 produces answers with an average of 3.1 steps, while the annotated answers have an average of 4.2 steps, which further confirms this observation.

\subsection{Analysis and Discussions}
\label{analysis}

In this section, we delve into two key questions:
(1) the alignment between our metric and human judgment;
(2) the underlying causes of VLMs' poor performance.

\begin{figure}[t]
  \centering
   \includegraphics[width=\linewidth, trim=0.4cm 2.2cm 0.8cm 0.4cm, clip]{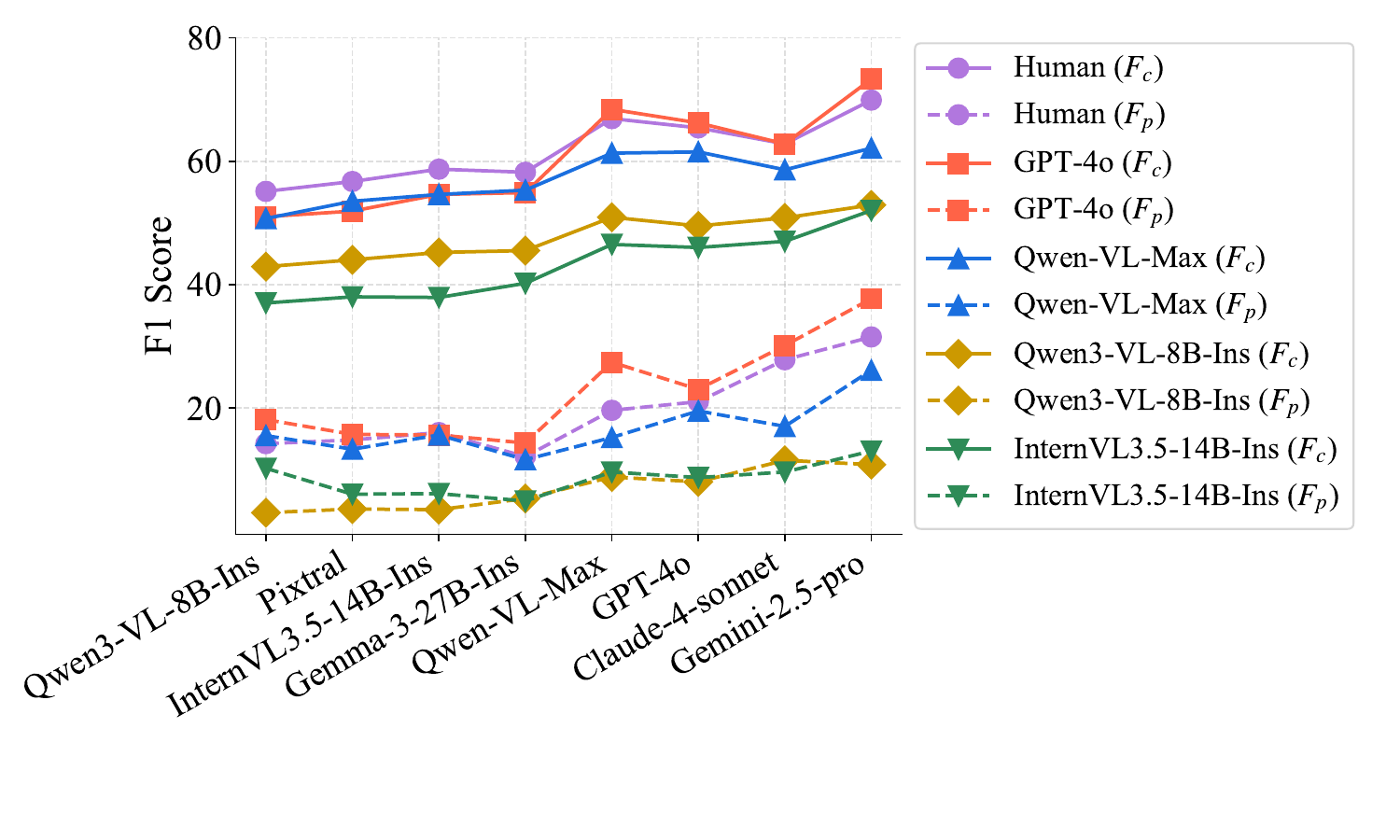}
   \vspace{-2.0em}
   \caption{Evaluation results of human and different scoring VLMs. The x-axis represents the eight representative VLMs being evaluated. Each point with the same marker shape denotes the F1 scores obtained by the same scoring VLM or by human evaluators. The solid lines indicate the F1 scores for content, while the dashed lines represent the F1 scores for preconditions.}
   \vspace{-1.5em}
   \label{fig:matrix_model}
\end{figure}

\paragraph{Human Alignment of VLM-based Evaluation.}
We employ GPT-4o when matching predicted steps with annotated ones. In the following analysis, we investigate the consistency between the \textit{scoring} VLM (used to generate the matching matrix) and the human evaluators.
To this end, we selected eight representative VLMs as the evaluated models and four VLMs as the scoring VLMs. All evaluations were conducted on 300 randomly sampled instances from SpatiaLQA. \cref{fig:matrix_model} presents the evaluation results obtained from human evaluators and different scoring VLMs, while \cref{tab:vs_human} compares the outcomes between the scoring VLMs and human evaluators.
The results show that evaluation scores vary significantly depending on which VLM is used as the scoring model. Notably, proprietary models (Qwen-VL-Max and GPT-4o) yield results that are more consistent with human evaluations, likely because they learn more stable semantic similarity judgment patterns from larger, higher-quality data, bringing their judgments closer to human intuition. Specifically, proprietary models generally exhibit higher correlation coefficients and lower mean absolute errors (around 3 percentage points), whereas open-source models show mean absolute errors exceeding 10 percentage points. Furthermore, GPT-4o achieves the highest correlation and lowest mean absolute error.
Therefore, to maintain consistency with human judgment, we adopt GPT-4o as the scoring VLM.

\begin{figure}[t]
  \centering
   \includegraphics[width=\linewidth, trim=11.1cm 12.4cm 11.4cm 1.0cm, clip]{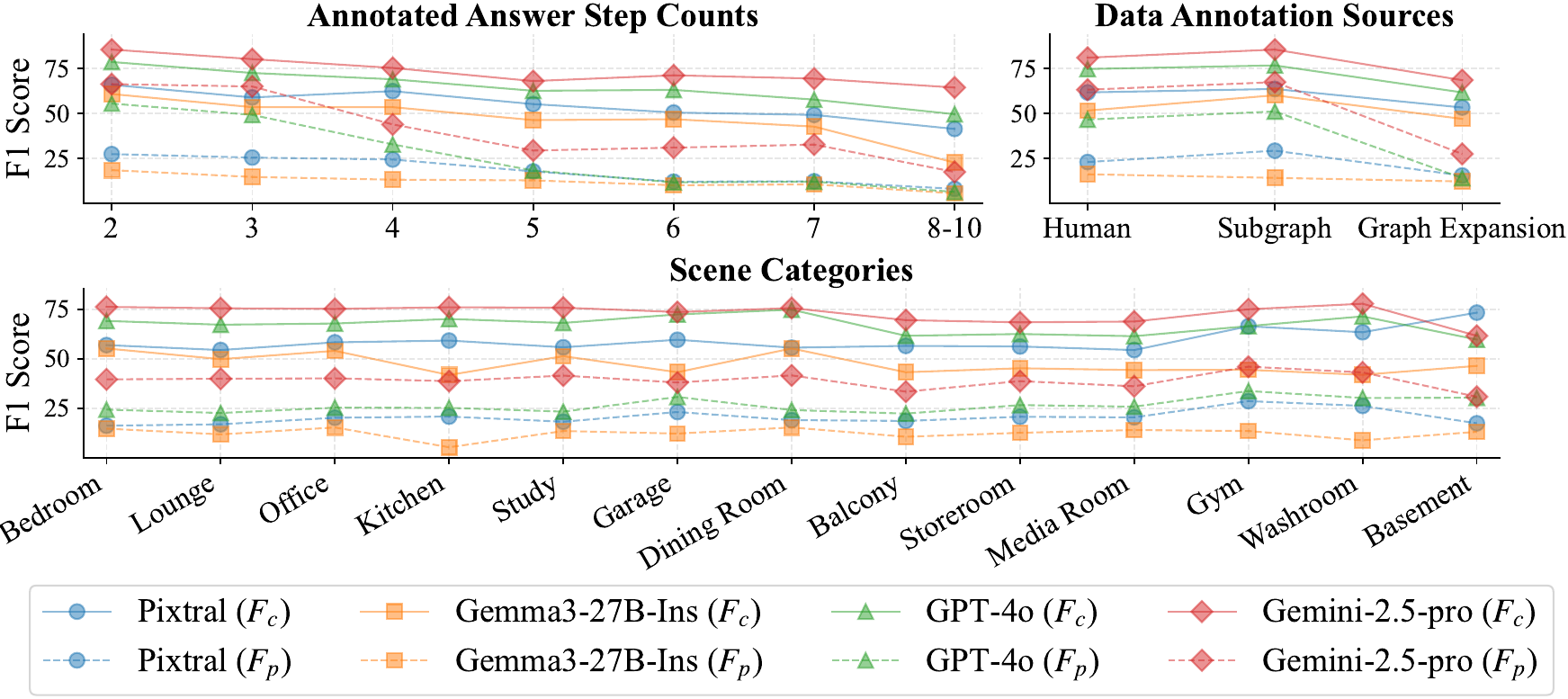}
   \vspace{-2.0em}
   \caption{Dimension-wise analysis. Each dot represents the F1 score of a VLM under specific conditions, including different numbers of answer steps, annotation methods, and scene categories. `Human', `Subgraph', and `Graph Expansion' correspond to data from `manual annotations', `subgraph extraction augmentation', and `graph expansion augmentation', respectively.}
   \vspace{-1.5em}
   \label{fig:analysis}
\end{figure}

\paragraph{The Underlying Causes of Poor Performance.}
To answer this question, we selected four representative VLMs and analyzed their performance more comprehensively from three dimensions: the number of annotated answer steps, annotation source, and scene category. As shown in \cref{fig:analysis}, model performance generally decreases as the number of annotated answer steps increases. Moreover, model performance shows clear patterns across different annotation sources: VLMs perform best on data generated by subgraph extraction augmentation, followed by manual annotations, and worst on graph expansion augmentation. This trend arises because the samples generated through subgraph extraction augmentation have fewer answer steps (simpler problems), whereas the samples generated through graph expansion augmentation contain more steps (more complex problems). However, VLMs exhibit relatively consistent performance across various scene categories.

These observations suggest that VLMs tend to perform worse on tasks requiring more steps, as such tasks demand longer and more stable reasoning processes. VLMs' failures on these tasks result in overall poor performance.

\begin{table}
\small
\centering
\begin{tabular}{ccccc}

\Xhline{1.0pt}
 \textbf{Scoring VLMs}  & $\rho_c$ & $\rho_p$ & $s$ & $\Delta$ \\ \hline

 Qwen3-VL-8B-Ins & 0.96 & 0.89 & 0.98 & 13.4 \\
 InternVL3.5-14B-Ins & 0.96 & 0.78 & 0.99 & 14.9 \\
 Qwen-VL-Max & 0.97 & 0.86 & 0.99 & 3.5 \\
 GPT-4o & 0.99 & 0.96 & 0.99 & 3.0 \\
  \Xhline{1.0pt}
\end{tabular}
\vspace{-1em}
\caption{Comparison between scoring VLM evaluation results and human evaluation results. $\rho_c$ ($\rho_p$) represents the Pearson correlation coefficient between the content (precondition) F1 scores obtained by VLMs and those obtained by humans. $s$ ($\Delta$) denotes the cosine similarity (mean absolute error) between the F1 scores obtained by VLMs and those obtained by humans.}
\label{tab:vs_human}
\vspace{-1.em}
\end{table}

\section{Recursive Scene Graph Assisted Reasoning}
\label{sec:method}

To address the poor performance of VLMs on complex tasks, we introduce Recursive Scene Graph Assisted Reasoning (RSGAR) in \cref{method} and present its effectiveness and ablation studies in \cref{exp}. Please refer to the supplementary material for details on hyperparameters and costs.

\subsection{Method}
\label{method}

\begin{figure}[t]
  \centering
   \includegraphics[width=\linewidth, trim=0.7cm 0.9cm 0.7cm 0.9cm, clip]{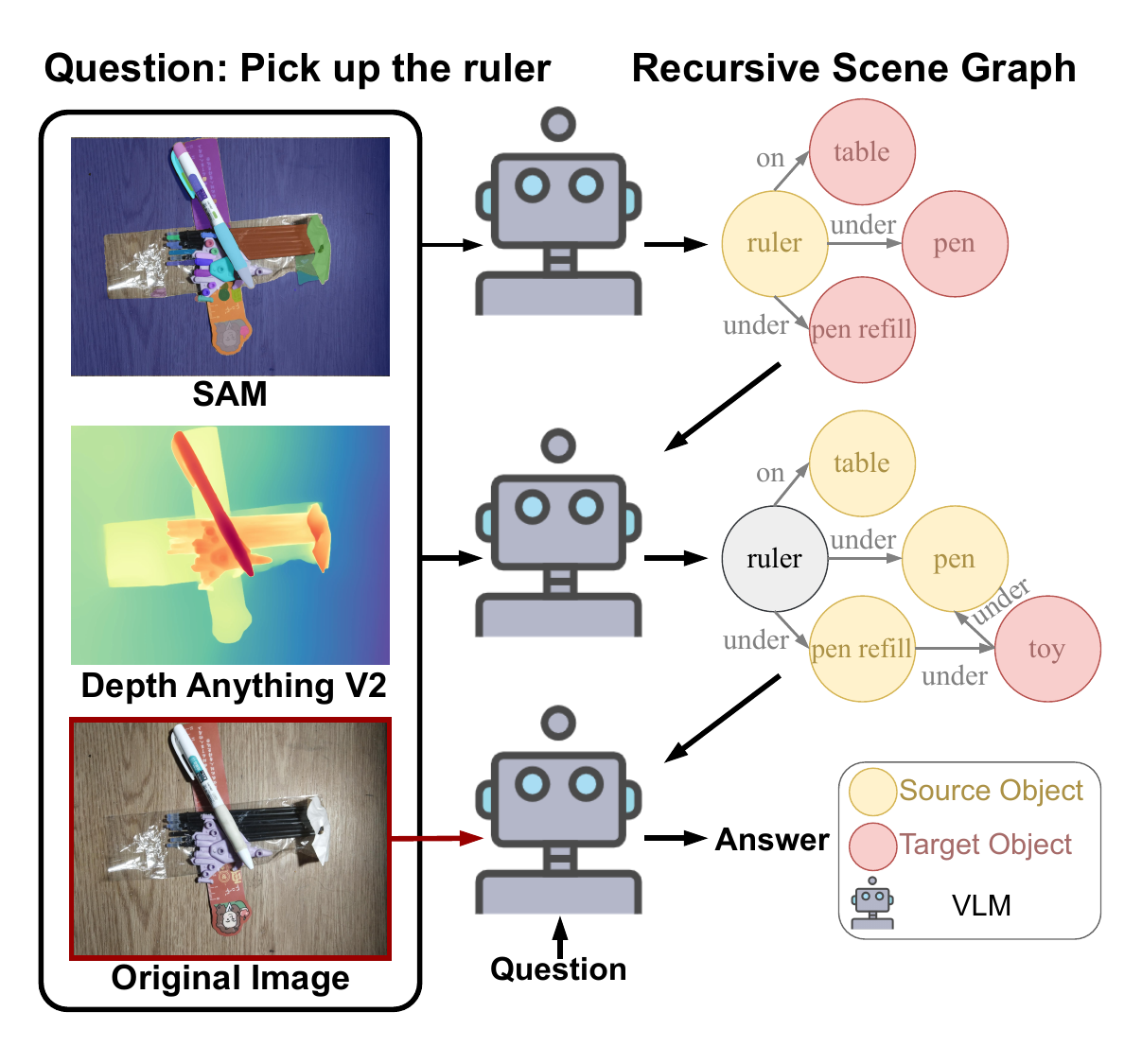}
   \vspace{-2.0em}
   \caption{The overview of RSGAR. Scene graph generation and question answering are performed by the same VLM.}
   \vspace{-1.5em}
   \label{fig:method}
\end{figure}

As shown in \cref{fig:method}, RSGAR consists of the following three steps:
(1) We first employ Depth Anything V2 and SAM to obtain the depth and segmentation maps of the scene image.
(2) Using these perception results together with the original image, we designate the task-specified target as the initial source object and perform the first round of scene graph generation with the VLM. During this process, the model identifies the objects in direct contact with the source object, referred to as target objects, and determines their spatial relationships. A scene graph is then constructed with the source and target objects as nodes and their spatial relationships as edges. This graph serves as input for the next iteration, where the previous target objects are treated as new source objects. The process continues until a predefined maximum iteration $T$ is reached.
(3) Finally, the generated scene graph is combined with the task prompt and fed into the VLM to produce the final answer.

\subsection{Experiments}
\label{exp}

\paragraph{Effectiveness.} We adopt five baselines: PhysAgent~\cite{chow2025physbench}, Chain of Thought (CoT)~\cite{wei2022chain}, and three variants of vanilla reasoning with additional inputs of segmentation map (SAM), depth map (Depth Anything V2), or both together. PhysAgent is one of the most advanced methods for enhancing VLMs’ physical commonsense and understanding of the real world. We use GPT-4o as the VLM for all methods, and the number of iterations $T$ in RSGAR is set to 5. The results of all methods on SpatiaLQA are shown in \cref{tab:main_exp}, which shows that RSGAR achieves the best performance among all methods, while CoT attains the second-best result due to its ability to enable more stable reasoning. Other baselines even underperform the vanilla reasoning, indicating that simply incorporating physical priors or visual cues does not contribute to the spatial logical reasoning of VLMs.

\paragraph{The Reason for Effectiveness.} To answer this question, we further analyzed RSGAR from the perspective of the number of steps in the annotated answers. As shown in \cref{tab:exp_step}, although RSGAR shows a slight decrease in performance on samples with fewer answer steps, it significantly improves the performance of VLMs on samples with more answer steps. This suggests that RSGAR can improve VLM performance on complex tasks by explicitly representing the relationships among key objects in the original scene.

\begin{table}
\centering
\small
\addtolength{\tabcolsep}{-4.0pt}
\begin{minipage}{0.18\textwidth}
\centering
\begin{tabular}{ccc}
\Xhline{1.0pt}
 &\textbf{$F_c$} &\textbf{$F_p$} \\ \hline
 GPT-4o & 67.4 & 25.1 \\
 + depth & 64.1 & 19.0 \\
 + seg & 50.3 & 14.7 \\
 + seg\&depth & 50.5 & 14.5 \\
 PhysAgent & 64.7 & 22.2 \\
 CoT & \underline{67.6} & \underline{27.0} \\
 RSGAR & \textbf{69.8} & \textbf{28.1} \\
  \Xhline{1.0pt}
\end{tabular}
\subcaption{RSGAR vs. baselines.}
\label{tab:main_exp}
\end{minipage}
\hfill \hspace{2.5pt}
\begin{minipage}{0.28\textwidth}
\centering
\begin{tabular}{ccc}
\Xhline{1.0pt}
 Step Counts&\textbf{$F_c$} &\textbf{$F_p$} \\ \hline
 2 & 76.4 (-1.3) & 53.9 (-0.7) \\
 3 & 71.6 (-1.0) & 46.4 (-1.9) \\
 4 & 73.6 \green{(+4.5)} & 34.8 \green{(+1.9)} \\
 5 & 65.6 \green{(+2.8)} & 20.0 \green{(+1.6)} \\
 6 & 62.8 (-0.4) & 14.3 \green{(+2.5)} \\
 7 & 58.6 \green{(+0.8)} & 15.2 \green{(+3.0)} \\
 8-10 & 50.2\green{(+0.5)} & 8.9\green{(+2.4)} \\
  \Xhline{1.0pt}
\end{tabular}
\subcaption{$F_c$/$F_p$ across answer step counts.}
\label{tab:exp_step}
\end{minipage}
\vspace{-1em}
\caption{(a) The results of various methods. \textbf{Bold} and \underline{underlined} scores denote the best and second-best results. `+ depth', `+ seg', and `+ depth\&seg' represent vanilla reasoning enhanced with depth maps, segmentation maps, and both, respectively. (b) RSGAR's $F_c$ and $F_p$ across answer step counts. The values in parentheses represent the changes relative to vanilla reasoning.}
\vspace{-1.em}
\end{table}

\begin{table}
\centering
\small
\addtolength{\tabcolsep}{1.5pt}
\begin{minipage}{0.18\textwidth}
\centering
\begin{tabular}{x{12}x{16}x{16}}
\Xhline{1.0pt}
$T$ & $F_c$ & $F_p$ \\
\hline
1 & 68.5 & 27.4 \\
3 & 68.8 & 28.1 \\
5 & \cellcolor{gray!25} \textit{69.8} & \cellcolor{gray!25} \textit{28.1} \\
7 & 70.6 & 28.7 \\
  \Xhline{1.0pt}
\end{tabular}
\subcaption{Iteration number $T$.}
\label{tab:exp_T}
\end{minipage}
\hfill
\hspace{2.5pt}
\begin{minipage}{0.28\textwidth}
\centering
\begin{tabular}{x{60}x{16}x{16}}
\Xhline{1.0pt}
 & $F_c$ & $F_p$ \\
\hline
w/o seg\&depth & 66.5 & 26.3 \\
w/o seg & 66.9 & 26.5 \\
w/o depth & 68.8 & 27.8 \\
w/ seg\&depth & \cellcolor{gray!25} \textit{69.8} & \cellcolor{gray!25} \textit{28.1} \\
  \Xhline{1.0pt}
\end{tabular}
\subcaption{Segmentation/Depth map.}
\label{tab:exp_map}
\end{minipage}
\vspace{-1em}
\caption{Ablation studies on GPT-4o. The default settings are indicated in \textit{italics}. `seg' and `depth' denote the segmentation map and depth map, respectively.}
\label{tab:exp_ablation}
\vspace{-1.5em}
\end{table}

\paragraph{Ablation Studies.} We set \( T = 5 \) by default and provide both the depth map and segmentation map during scene graph generation, as indicated by the \textit{italics} in \cref{tab:exp_ablation}. \cref{tab:exp_T} shows that the performance of the VLM improves as \( T \) increases. This is because a larger \( T \) allows the final scene graph to include more information, giving the VLM a more comprehensive understanding of the scene when answering questions. \cref{tab:exp_map} shows that the performance of RSGAR declines when either the depth map or segmentation map is not used. This is because each of them provides distinct visual information, and both are essential for generating an accurate scene graph.

\section{Conclusion}
\label{sec:conclusion}

\begin{spacing}{1.00}
In summary, we introduce SpatiaLQA, a benchmark that systematically evaluates the spatial logical reasoning of VLMs.
Moreover, we develop an automated evaluation strategy based on GPT-4o, which achieves a high level of consistency with human.
By evaluating 41 VLMs, we reveal significant deficiencies in their spatial logical reasoning.
Therefore, we propose recursive scene graph assisted reasoning, which effectively enhances GPT-4o's performance on complex tasks.
We will further explore how VLMs can be applied to more complex scenarios.
\end{spacing}
{
    \small
    \bibliographystyle{ieeenat_fullname}
    \bibliography{main}
}

\clearpage
\setcounter{page}{1}
\maketitlesupplementary
\appendix
\setcounter{table}{0}
\renewcommand{\thetable}{A\arabic{table}}

\section{SpatiaLQA Benchmark Details}
\label{app:bench_details}

This section provides additional details on the construction of the SpatiaLQA benchmark. Specifically, \cref{app:rules} describes the data collection rules, and \cref{app:manual} outlines the manual annotation and review process.

\subsection{Data Collection Rules}
\label{app:rules}

During data collection, we followed the rules below to ensure both diversity and category balance in the dataset:

(1) The same object is allowed to appear no more than ten times within a single scene, and its appearance frequency should be kept as low as possible across the entire dataset.

(2) The same scene (\eg, the same desk, the same sofa) may appear approximately 20 times. Changes in camera angle, lighting, \etc, do not count as a new scene.

(3) The scene layout may be modified without changing the scene itself. For example, using an office desk as the base scene, one may rearrange or replace the items on the desk (each item still appearing no more than ten times).

\subsection{Manual Annotation and Review}
\label{app:manual}

The SpatiaLQA dataset consists of an image directory and a JSON file.
The image directory contains multiple subfolders, each corresponding to a specific scene. Subfolders follow the naming format `sceneIndex-sceneName', such as `0001-office-1' or `0022-bedroom-2'.
The prefix number indicates the global index of the scene across the entire dataset, while the suffix number denotes the index of that scene type (\eg, bedroom-2 refers to the folder containing all images of the second bedroom scene).

Each sample annotation consists of four components: the question, the answer, the corresponding image path, and the associated scene category. After all annotations are completed, they are reviewed by designated annotators. During review, each sample is examined for step validity and prerequisite correctness, with outcomes marked as either `approved' or `rejected, with reasons provided.'
The annotation–review cycle is repeated twice to ensure that all annotations are accurate and logically sound.

\section{Evaluation Details}
\label{app:eval_details}

In this section, we present the prompts, hyperparameters, and other settings used in our evaluation. Specifically, \cref{app:hyper_matrix} describes the prompts and hyperparameters used when generating the matching matrices with GPT-4o, and \cref{app:hyper_eval} details the prompts and hyperparameters used for evaluating the various VLMs.

\subsection{The Details of Generating the Matching Matrices}
\label{app:hyper_matrix}

The prompt used to generate the matching matrix with GPT-4o is as follows:

\begin{tcolorbox}[colback=gray!15, colframe=gray!60]
You are given an image and two list of step descriptions (Ground Truth List and Predicted List), each containing multiple sentences describing actions.

\vspace{\baselineskip}

Your task is to compare each sentence in Ground Truth List with each sentence in Predicted List, and determine whether they describe the same action in the context of the image. Two sentences are considered the same if they refer to the same objects and the same operations with no significant difference in meaning or execution.

\vspace{\baselineskip}

Please output an \verb|m × n| matrix where:

1. \verb|m| is the number of steps in Ground Truth List

2. \verb|n| is the number of steps in Predicted List

3. Each cell \verb|[i][j]| is either 1 (Yes) or 0 (No), indicating whether 
sentence \verb|i| from Ground Truth List is equivalent in meaning to sentence \verb|j| from Predicted List

\vspace{\baselineskip}

Ground Truth List:
\{ground\_truth\_steps\}

Predicted List:
\{predicted\_steps\}

\vspace{\baselineskip}

Please output an \verb|m × n| matrix, with each element being either 1 or 0. You only need to return a list of lists wrapped with \verb|<ans></ans>| tags (\eg, \verb|<ans>[[0, 1], [1, 1]]</ans>|).
\end{tcolorbox}

where \{ground\_truth\_steps\} and \{predicted\_steps\} denote the annotation result and the VLM’s prediction, respectively. Regarding hyperparameters, we set the temperature to 0.

\subsection{The Details of Evaluating the Various VLMs}
\label{app:hyper_eval}

For open-source VLMs, we use the corresponding HuggingFace models for local inference. The version of transformers used matches that of VLMEvalKit. For models not covered by VLMEvalKit~\cite{duan2024vlmevalkit}, we adopt the model versions recommended on HuggingFace. In addition, all hyperparameters for open-source models follow the recommended settings provided on HuggingFace.
For proprietar VLMs, we uniformly set the temperature to 0.

The prompt used for evaluating the VLMs is as follows:

\begin{tcolorbox}[colback=gray!15, colframe=gray!60]
Given a scene image and a task, please provide a concise and accurate list of steps to execute the task in order.

\vspace{\baselineskip}

Here is an example:
\{example\}

\vspace{\baselineskip}

Each step in the answer consists of two parts:
the action to be performed in this step (`content') and the steps that must be completed beforehand (`precondition').

\vspace{\baselineskip}

Please note the following:

1. The example above is only for illustrating the answer format and is not related to the question you need to answer.

2. All objects in the image are within your reach, so no movement is necessary.

3. Each step can only perform one action on a single object.

4. If you need to remove an object, use the action `remove' directly.

5. Write your answer in the format above (the answer consists of multiple steps, and each step contains a `content' and a `precondition'), enclosed within \verb|<ans></ans>| tags.

\vspace{\baselineskip}

Now answer the following question based on the input image: \{question\}. You only need to return an answer wrapped with \verb|<ans></ans>| tags, and the answer should be in JSON format (as shown in the `answer' field of the example).
\end{tcolorbox}

Here, `\{example\}' is a JSON-formatted sample, as shown below (similarly for the rest){}:

\begin{lstlisting}[language=json]
{
    "question": "Pick up the laptop",
    "answer": {
        "step1":{
            "content": "Remove the stapler from the top of the book",
            "precondition":[]
        }, 
        "step2":{
            "content": "Remove the keys from the top of the book",
            "precondition":[]
        }, 
        "step3":{
            "content": "Remove the toliet paper from the top of the laptop",
            "precondition":[]
        }, 
        "step4":{
            "content": "Remove the book from the top of the laptop",
            "precondition":["step1", "step2"]
        }, 
        "step5":{
            "content": "Pick up the laptop",
            "precondition":["step3", "step4"]
        }
    }
}
\end{lstlisting}

\section{The Details of the Baselines and RSGAR}
\label{app:exp}

For both the baselines and our proposed method RSGAR, the temperature of GPT-4o is fixed at 0. In this section, we also present the prompts used by the baselines and by RSGAR. Specifically, \cref{app:baseline} and \cref{app:RSGAR} describe the prompts for the baselines and for RSGAR, respectively.

\subsection{The Details of Baselines}
\label{app:baseline}

The prompt used for `+ depth’ is as follows:

\begin{tcolorbox}[colback=gray!15, colframe=gray!60, breakable]
Given a scene image, the depth map of the image and a task, please provide a concise and accurate list of steps to execute the task in order.

\vspace{\baselineskip}

Here is an example:
\{example\}

\vspace{\baselineskip}

Each step in the answer consists of two parts:
the action to be performed in this step (`content') and the steps that must be completed beforehand (`precondition').

\vspace{\baselineskip}

Please note the following:

1. The example above is only for illustrating the answer format and is not related to the question you need to answer.

2. All objects in the image are within your reach, so no movement is necessary.

3. Each step can only perform one action on a single object.

4. If you need to remove an object, use the action `remove' directly.

5. Write your answer in the format above (the answer consists of multiple steps, and each step contains a `content' and a `precondition'), enclosed within \verb|<ans></ans>| tags.

\vspace{\baselineskip}

Now answer the following question based on the input image: \{question\}. You only need to return an answer wrapped with \verb|<ans></ans>| tags, and the answer should be in JSON format (as shown in the `answer' field of the example).
\end{tcolorbox}

The prompt used for `+ seg’ is as follows:

\begin{tcolorbox}[colback=gray!15, colframe=gray!60, breakable]
Given a scene image, the segmentation map of the image and a task, please provide a concise and accurate list of steps to execute the task in order.

\vspace{\baselineskip}

Here is an example:
\{example\}

\vspace{\baselineskip}

Each step in the answer consists of two parts:
the action to be performed in this step (`content') and the steps that must be completed beforehand (`precondition').

\vspace{\baselineskip}

Please note the following:

1. The example above is only for illustrating the answer format and is not related to the question you need to answer.

2. All objects in the image are within your reach, so no movement is necessary.

3. Each step can only perform one action on a single object.

4. If you need to remove an object, use the action `remove' directly.

5. Write your answer in the format above (the answer consists of multiple steps, and each step contains a `content' and a `precondition'), enclosed within \verb|<ans></ans>| tags.

\vspace{\baselineskip}

Now answer the following question based on the input image: \{question\}. You only need to return an answer wrapped with \verb|<ans></ans>| tags, and the answer should be in JSON format (as shown in the `answer' field of the example).
\end{tcolorbox}

The prompt used for `+ seg\&depth’ is as follows:

\begin{tcolorbox}[colback=gray!15, colframe=gray!60, breakable]
Given a scene image, the depth map of the image, the segmentation map of the image and a task, please provide a concise and accurate list of steps to execute the task in order.

\vspace{\baselineskip}

Here is an example:
\{example\}

\vspace{\baselineskip}

Each step in the answer consists of two parts:
the action to be performed in this step (`content') and the steps that must be completed beforehand (`precondition').

\vspace{\baselineskip}

Please note the following:

1. The example above is only for illustrating the answer format and is not related to the question you need to answer.

2. All objects in the image are within your reach, so no movement is necessary.

3. Each step can only perform one action on a single object.

4. If you need to remove an object, use the action `remove' directly.

5. Write your answer in the format above (the answer consists of multiple steps, and each step contains a `content' and a `precondition'), enclosed within \verb|<ans></ans>| tags.

\vspace{\baselineskip}

Now answer the following question based on the input image: \{question\}. You only need to return an answer wrapped with \verb|<ans></ans>| tags, and the answer should be in JSON format (as shown in the `answer' field of the example).
\end{tcolorbox}

The prompt used for `CoT’ is as follows:

\begin{tcolorbox}[colback=gray!15, colframe=gray!60, breakable]
Given a scene image and a task, please provide a concise and accurate list of steps to execute the task in order.

\vspace{\baselineskip}

Here is an example:
\{example\}

\vspace{\baselineskip}

Each step in the answer consists of two parts:
the action to be performed in this step (`content') and the steps that must be completed beforehand (`precondition').

\vspace{\baselineskip}

Please note the following:

1. The example above is only for illustrating the answer format and is not related to the question you need to answer.

2. All objects in the image are within your reach, so no movement is necessary.

3. Each step can only perform one action on a single object.

4. If you need to remove an object, use the action `remove' directly.

5. Write your answer in the format above (the answer consists of multiple steps, and each step contains a `content' and a `precondition'), enclosed within \verb|<ans></ans>| tags.

\vspace{\baselineskip}

You should first think step by step to reason about the task, identifying all relevant objects, dependencies, and the logical order of actions. After finishing your reasoning, summarize only the final structured answer in JSON format, wrapped with \verb|<ans></ans>| tags, without including your intermediate reasoning in the output.

Now answer the following question based on the input image: \{question\}. You only need to return an answer wrapped with \verb|<ans></ans>| tags, and the answer should be in JSON format (as shown in the `answer' field of the example).
\end{tcolorbox}

For PhysAgent~\cite{chow2025physbench}, we follow the inference procedure described in the original paper, and all hyperparameters are kept consistent with the original settings.

\subsection{The Details of RSGAR}
\label{app:RSGAR}

The prompt used in the first round of scene-graph generation is as follows:

\begin{tcolorbox}[colback=gray!15, colframe=gray!60, breakable]
You are given an image (along with its depth map and segmentation map) and a natural language task instruction. You have two tasks, \textbf{Target Object Identification} and \textbf{Scene Graph Generation}

\vspace{\baselineskip}

\textbf{1. Target Object Identification}

Based on the task instruction, identify the source objects that are directly involved in accomplishing the task.
For example:

a. `Pour a glass of water' → both the kettle (or teapot) and the glass are source objects.

b. `Pick up the teapot' → the teapot is a source object.

\vspace{\baselineskip}

\textbf{2. Scene Graph Generation}

Based on the input image and source objects, build a scene graph centered on the source objects.
For each source object, list:

a. Its category/name.

b. Include only target objects that have a direct spatial relationship (direct contact or adjacency) or a direct functional relationship with the source object.

\vspace{\baselineskip}

Here is an example:
\{example\}

\vspace{\baselineskip}

The example above is only for illustrating the answer format and is not related to the question you need to answer. Ensure that the `source' in the scene graph is exclusively the source object. `relation' is not limited to the given example and can be any word.

\vspace{\baselineskip}

Now finish the \textbf{Target Object Identification} and \textbf{Scene Graph Generation} based on the original image (maps for reference only) and question `\{question\}'. You only need to return an answer wrapped with \verb|<ans></ans>| tags.
\end{tcolorbox}

Here, `\{example\}' is a JSON-formatted sample, as shown below (similarly for the rest):

\begin{lstlisting}[language=json]
{
    "source_objects": [
        {
            "name": "teapot",
            "attributes": ["silver", "full of water", "lid closed"],
            "reason": "Required to pour water"
        },
        {
            "name": "cup",
            "attributes": ["ceramic", "empty", "handle on right side"],
            "reason": "Receives water"
        }
    ],
    "scene_graph": [
        {"source": "teapot", "relation": "on", "target": "tray"},
        {"source": "cup", "relation": "next to", "target": "teapot"},
        {"source": "teapot", "relation": "under", "target": "box"}
    ]
}
\end{lstlisting}

The prompt used for generating the scene graph after the first round is as follows:

\begin{tcolorbox}[colback=gray!15, colframe=gray!60, breakable]
Given an image (along with its depth map and segmentation map) and several source objects with their scene graphs in history outputs, you are required to take the target objects (`target') in the scene graph of history outputs as the new source objects, and generate a new scene graph based on these source objects.

\vspace{\baselineskip}

Here is the history outputs:
\{history\_outputs\}

\vspace{\baselineskip}

For each source object in this stage, list:

a. Its category/name.

b. Only the target objects that have a direct spatial relationship (direct contact or adjacency) or a direct functional relationship with the new source object.

\vspace{\baselineskip}

Here is an example:
\{example\}

\vspace{\baselineskip}

The example above is only for illustrating the answer format and is not related to the question you need to answer. Ensure that the `source' in the scene graph is exclusively the new source object. `relation' is not limited to the given example and can be any word. `target' must be something that has not appeared in the scene graph of history outputs.

\vspace{\baselineskip}

Finally, you only need to output the source objects and their scene graphs for this stage, following the format of the example. The answer should be wrapped with \verb|<ans></ans>| tags.
\end{tcolorbox}

Here, `\{history\_outputs\}' refers to the scene graphs generated in the previous rounds.

The prompt used when incorporating the scene graph for assisted reasoning is as follows:

\begin{tcolorbox}[colback=gray!15, colframe=gray!60, breakable]
Given a scene image (along with its scene graph, where the scene graph depicts spatial relationships among certain objects in the image) and a task, please provide a concise and accurate list of steps to execute the task in order.

\vspace{\baselineskip}

Here is the scene graph of the input image:
\{scene\_graph\}

\vspace{\baselineskip}

Here is an answer example:
\{example\}

\vspace{\baselineskip}

Each step in the answer consists of two parts:
the action to be performed in this step (`content') and the steps that must be completed beforehand (`precondition').

\vspace{\baselineskip}

Please note the following:

1. The example above is only for illustrating the answer format and is not related to the question you need to answer.

2. All objects in the image are within your reach, so no movement is necessary.

3. Each step can only perform one action on a single object.

4. If you need to remove an object, use the action `remove' directly.

5. Write your answer in the format above (the answer consists of multiple steps, and each step contains a `content' and a `precondition'), enclosed within \verb|<ans></ans>| tags.

\vspace{\baselineskip}

Now answer the following question based on the input image and scene graph (the scene graph may not be accurate and is for reference only): \{question\}. You only need to return an answer wrapped with \verb|<ans></ans>| tags, and the answer should be in JSON format (as shown in the `answer' field of the example).
\end{tcolorbox}

Here, `\{scene\_graph\}' refers to the previously generated scene graph.

\section{Efficiency Analysis}
\label{app:effi}

We measured the time required for RSGAR and the other baselines to perform a single round of reasoning over the entire SpatiaLQA dataset. All experiments were conducted on an NVIDIA GeForce RTX 4090 and the VLM used was GPT-4o. The prompts and hyperparameters for each method are provided in \cref{app:exp}. The results are shown below:

\begin{table}[ht]
\centering
\small
\begin{tabular}{cccc}
\Xhline{1.0pt}
Method & $\tau$ & $F_c$ & $F_p$\\ \hline
Vanilla Reasoning & 27.4h & 67.4 & 25.1 \\
+ depth & 27.9h & 64.1 & 19.0 \\
+ seg & 29.4h & 50.3 & 14.7 \\
+ seg\&depth & 30.1h & 50.5 & 14.5 \\
PhysAgent & 31.5h & 64.7 & 22.2 \\
CoT & 27.5h & 67.6 & 27.0 \\
RSGAR ($T=1$) & 57.9h & 68.5 & 27.4 \\
RSGAR ($T=5$) & 174.5h & 69.8 & 28.1 \\
\Xhline{1.0pt}
\end{tabular}
\caption{Efficiency analysis. \( \tau \) is the time taken by each method to perform one validation on SpatiaLQA.}
\label{tab:time}
\vspace{-1.5em}
\end{table}

The results in \cref{tab:time} show that although RSGAR requires longer inference time, it achieves better performance than the other baselines. This is because RSGAR is task-oriented and progressively transforms the original image into an interpretable scene graph, providing the VLM with additional information during the reasoning phase, thereby improving overall performance.

\end{document}